\documentclass{article}
\usepackage{arxiv}
\pdfoutput=1

\usepackage[utf8]{inputenc} 
\usepackage[T1]{fontenc}    
\usepackage[bookmarks=false]{hyperref}      
\usepackage{url}            
\usepackage{booktabs}       
\usepackage{amsfonts}       
\usepackage{nicefrac}       
\usepackage{microtype}      
\usepackage{lipsum}		
\usepackage{graphicx}
\usepackage{multirow}
\usepackage{natbib}
\usepackage{doi}

\title{BERTuit: Understanding Spanish language in Twitter through a native transformer}

\author{ \href{https://scholar.google.es/citations?user=5XOhXooAAAAJ}{\includegraphics[scale=0.02]{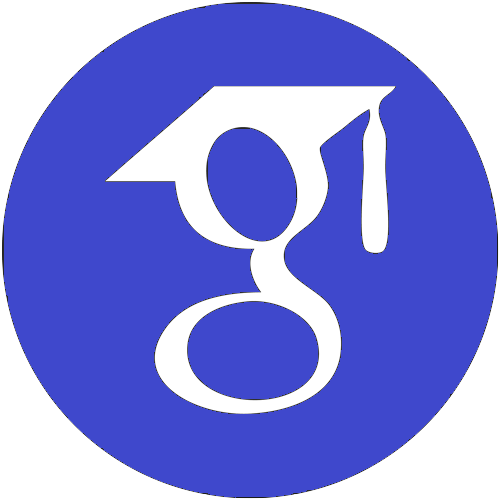}\hspace{1mm}Javier Huertas-Tato} \\
	Departamento de Sistemas Informáticos\\
	Universidad Politécnica de Madrid\\
	Spain, Madrid 28031 \\
	\texttt{javier.huertas.tato@upm.es} \\
	\And
	\href{https://scholar.google.com/citations?user=b3J9VRsAAAAJ}{\includegraphics[scale=0.02]{g_scholar.png}\hspace{1mm}Alejandro Martín} \\
	Departamento de Sistemas Informáticos\\
	Universidad Politécnica de Madrid\\
	Spain, Madrid 28031 \\
	\texttt{alejandro.martin@upm.es} \\
	\And
	\href{https://scholar.google.com/citations?user=fpf6EDAAAAAJ}{\includegraphics[scale=0.02]{g_scholar.png}\hspace{1mm}David Camacho} \\
	Departamento de Sistemas Informáticos\\
	Universidad Politécnica de Madrid\\
	Spain, Madrid 28031 \\
	\texttt{david.camacho@upm.es} \\
}

\renewcommand{\shorttitle}{\textit{arXiv} Template}

\hypersetup{
pdftitle={BERTuit: Understanding Spanish language in Twitter through a native transformer},
pdfsubject={cs.CL},
pdfauthor={Javier Huertas-Tato, Alejandro Martin, David Camacho},
pdfkeywords={Transformers, Online Social Networks, Twitter, Misinformation},
}

\begin{document}
\maketitle

\begin{abstract}

    The appearance of complex attention-based language models such as BERT, RoBERTa or GPT-3 has allowed to address highly complex tasks in a plethora of scenarios. However, when applied to specific domains, these models encounter considerable difficulties. This is the case of Social Networks such as Twitter, an ever-changing stream of information written with informal and complex language, where each message requires careful evaluation to be understood even by humans given the important role that context plays. Addressing tasks in this domain through Natural Language Processing involves severe challenges. When powerful state-of-the-art multilingual language models are applied to this scenario, language specific nuances get lost in translation. To face these challenges we present \textbf{BERTuit}, the largest transformer proposed so far for Spanish language, pre-trained on a massive dataset of 230M Spanish tweets using RoBERTa optimization. Our motivation is to provide a powerful resource to better understand Spanish Twitter and to be used on applications focused on this social network, with special emphasis on solutions devoted to tackle the spreading of misinformation in this platform.  BERTuit is evaluated on several tasks and compared against M-BERT, XLM-RoBERTa and XLM-T, very competitive multilingual transformers. The utility of our approach is shown with applications, in this case: a zero-shot methodology to visualize groups of hoaxes and profiling authors spreading disinformation. 
\end{abstract}

\keywords{Transformers \and Online Social Networks \and Twitter \and Misinformation}

\section{Introduction} \label{sec.intro}
Recent years have seen an explosion of available information. Online Social Networks (OSNs) produce text, video and images orders of magnitude faster than any human, let alone experts, can manage. In this environment, content sharing thrives, revealing trends on human relationships such as opinions, sentiments, political stances and so on. This availability of information also increases the likelihood of coming across information disorders, which have proven repeatedly to be a safety hazard. Misinformation has undermined trust in vaccines, fostered beliefs in ineffective (or even dangerous) pseudo-scientific therapies, and created disbelief in the effectiveness of public data-driven policy~\cite{larson2018biggest}. Understanding the phenomena of misinformation is crucial for public safety.

A possible avenue to understand misinformation on social media is to understand its users interactions through the text messages they publicly share. This can be done using Natural Language Processing (NLP) techniques, which remain a powerful approach to understand content delivered in OSNs~\cite{farzindar2015natural}. NLP is able to infer knowledge from stylistic and knowledge characteristics of raw text. Furthermore, the transformer architecture~\cite{vaswani2017attention} has meant a leap forward in NLP, meaning that new opportunities for understanding misinformation in social platforms have appeared. Social media language is quite different from formal datasets, containing short-hand, emoji, hashtags, grammatical errors, irony, leetspeak and other semantically-charged parts of speech that are missing on popular corpus, such as news articles, encyclopedias, books and journal publications. In contrast, most transformers use formal corpus, for example BERT~\cite{devlin2018bert} uses a corpus comprised of books, and Wikipedia; XLM-RoBERTa~\cite{conneau2019unsupervised} and T5~\cite{raffel2019exploring} rely on the Common Crawl\footnote{\href{https://commoncrawl.org/}{Common Crawl website and repository: https://commoncrawl.org/}}, which contains data from multiple sources where a small fraction of data belongs to OSNs, which may lead to a bias toward formal sources.

Another challenge against the characterization of misinformation is the language barrier. Numerous multi-language NLP models have been successfully built on the past years. On many domains they offer extreme flexibility and accuracy, reaching similar results to mono-lingual models. However, OSNs may pose challenges that general-purpose multi-lingual transformers may not be able to overcome. Texts published for social media relies on cultural context, irony and idioms to be interpreted, which are usually language-specific features of a speech that require some degree of cultural awareness.

Motivated by these issues, we present BERTuit, a transformer trained from scratch with text created by native speakers from Twitter. BERTuit has been trained with more than $230$ million Tweets from the Archive Twitter Stream Grab\footnote{\href{https://archive.org/details/twitterstream?tab=about}{Twitter Stream Grab About page: https://archive.org/details/twitterstream?tab=about}}, from 2021 to 2018. Using this massive amount of data and BERT-base architecture~\cite{devlin2018bert}, we replicate RoBERTa~\cite{liu2019roberta} optimization to perform self-supervised masked language modeling pre-training. The result is a transformer model that accurately inherits leanings, nuances and biases from Spanish Twitter, which later is useful applied to any downstream task in Twitter and informal scenarios, as well as misinformation understanding in particular. We performed an evaluation on several NLP tasks on Spanish Twitter, comparing against the current best alternative from the state-of-the-art, XLM-RoBERTa and alternatively multilingual BERT (M-BERT). In summary, the following contributions can be found in this paper:
\begin{itemize}
    \item A description of a transformer model that reliably outperforms state-of-the-art alternatives on Spanish Twitter problems.
    \item This transformer coupled with an appropriate methodology can enhance the understanding of misinformation on social media. We contribute methods to achieve this.
    \item A powerful approach to represent claims containing misinformation into a 2d space using embeddings from the proposed transformer.
    \item An assessment of the ability of BERTuit to extract relevant language patterns even from small sets of data in the context of author profiling on Twitter.
\end{itemize}

The remaining sections of this manuscript are organised as follows: Section~\ref{sec.sota} presents a description of the state-of-the-art literature, showing similar approaches, Section~\ref{sec.pretrain} describes the BERTuit model and the pre-training procedure, Section~\ref{sec.validation} provides a validation of BERTuit in multiple tasks and in comparison with state-of-the-art models, Section~\ref{sec.apps} shows two use cases of BERTuit that also provide interesting details of the performance and, finally, Section~\ref{sec.conclusions} presents a number of conclusions.

\section{Related work} \label{sec.sota}

\subsection{The Transformer architecture}

The Transformer architecture has been a turning point in addressing Human Language Understanding tasks. In contrast to previous approaches, the self-attention mechanism~\cite{vaswani2017attention} is an important step forward in the understanding of language, extracting deep and complex relations and information of the context and semantic. From its proposal several years ago, a plethora of architectures can be found in the literature, improving performance in many tasks or showing excellent skills in solving highly complex tasks such as question answering or text generation, among many others.

Due to the large size of these architectures, the most popular have been released as pre-trained models, to be later fine-tuned in order to undertake specific tasks. Thus, BERT~\cite{devlin2018bert} is one of the most popular pre-trained models for language understanding. This model, trained with Masked Language Modeling and Next Sentence Prediction tasks, has been used as the basis for implementing new models for specific tasks, such as language-specific models for Finnish~\cite{virtanen2019multilingual} or Spanish~\cite{canete2020spanish}, ligher versions such as DistilBERT~\cite{sanh2019distilbert} and applied to specific problems such as hate speech detection~\cite{mozafari2020hate}, sentiment analysis~\cite{singh2021sentiment}. 

Another important model widely used is RoBERTa~\cite{liu2019roberta}, using a similar encoder topology. The authors claimed that BERT was undertrained, and proposed a new training method which improves performances in comparison to BERT. The modifications included a larger training process, to remove the next sentence prediction objective, use of longer sequences and a dynamic use of the mask applied over the training data. As in the case of BERT, RoBERTa has been fine-tuned for specific languages such as dutch~\cite{delobelle2020robbert} or czech~\cite{straka2021robeczech} and for many specific problems such as metaphor identification~\cite{babieno2022miss}.

GPT~\cite{radford2018improving}, GPT-2~\cite{radford2019language} and GPT-3~\cite{brown2020language} are three of the most popular language models, integrating a decoder for text generation. The last one, GPT-3 was trained with a large a mount of data and involves 175 billion parameters, showing excellent performance in translation or question-answering tasks, among many others. Other models include BART~\cite{lewis2019bart}, a encoder-decoder architecture, specially designed for sequence-to-sequence tasks. It was trained with corrupted text with the goal of providing a correct output. T5~\cite{raffel2019exploring} employs a similar architecture, using a training process where each possible input is associated with a text in the output. Thus, different tasks are adopted to follow this training process.  More recently, XLM-R (XLM-RoBERTa) was proposed to improve mutilingual BERT, showing excellent results in low-resource languages in different tasks such as NLI tasks. 

Research lines on the Transformer architecture also includes attempts to deal with long sequences, such as the Memory Compressed Transformer~\cite{liu2018generating}. Different modifications towards achieving efficient models have also been proposed~\cite{tay2020efficient}, through the use of Fixed Patterns, Combination of Patterns, Learnable Patterns, Nueral Memory, Low-Rank Methods, Kernels, Recurrence, Downsampling, Sparse Models or Conditional Computation.

\subsection{Specialized transformer models} \label{sec.sota.specialized}

Limitations of general-purpose transformers on OSN text are shown by BERTweet~\cite{nguyen2020bertweet}, where the authors propose training BERT architecture from scratch using a corpus composed of Twitter text. This improvement at pre-training manages to outperform RoBERTa and XLM-RoBERTa on mono-lingual English tasks. Mono-lingual transformers can be adapted with language pairs to other languages, however multi-lingual models created with this method usually present deficiencies like Multilingual BERT~\cite{pires2019multilingual} (M-BERT). A viable alternative could be found in recent advances such as XLM-Twitter~\cite{barbieri2021xlm}, where a XLM-RoBERTa model is trained upon a multilingual corpora of twitter data. Although more powerful on twitter problems than XLM-RoBERTa, this model uses around >$10$ million tweets per language, meaning that mono-lingual understanding is limited to the amount of data present at training. Some features of similar languages can be generalized, but individual subtleties are never learned, or severely underfit.

The immediate solution to these problems is to train a transformer from scratch with a massive mono-lingual corpus composed of text from Twitter, as proposed in TWilBERT~\cite{gonzalez2021twilbert}. This transformer outperforms M-BERT on several tasks. However, as the authors point out in future works, it could still benefit from more data. Other languages have their specialized transformers for twitter such as ALBERTo~\cite{polignano2019alberto}, which is meant for Italian Twitter. Using the lessons from TWilBERT, our proposal focuses on masked language modeling and doubles the available data for pre-training to build a robust twitter mono-lingual model similar in essence to BERTweet.

\subsection{Transformers in the context of disinformation} \label{sec.sota.disinformation}

Many transformer-based solutions have been developed specializing in the topic of misinformation. For example, on EMET~\cite{schwarz2020emet} a custom encoder architecture featuring transformer blocks is used. It embeds texts to discover misinformation on Twitter. Although robust, no comparison is drawn against modern pre-trained transformers, indicating that further improvements could be reached with state-of-the-art methods. Others like exBAKE~\cite{jwa2019exbake} add extra data to BERT from news sources to understand information data. As good as it performs, this system runs into problems when faced with Twitter text. Both solutions specialize on misinformation detection and, while they are powerful on their own, they cannot match the strengths of pre-trained transformers. These models have two pressing flaws when applied to our domain: \textit{a)} data corpus are built with formal sources, while Twitter text is informal and, in many scenarios, vulgar; \textit{b)} there are language barriers that multi-lingual models cannot overcome, such as cultural subtext or idiomatic expressions. Models trained with mono-lingual corpora or have heavy biases towards a single language may experience difficulties capturing the aforementioned subtleties. These challenges compose on each other when tackled together, representing a serious obstacle to understand misinformation on OSNs. Therefore, we explore solutions that have previously been proposed in the literature to overcome said challenges.

\section{BERTuit Pre-training} \label{sec.pretrain}
Self-supervised learning is performed to pretrain our language model. BERTuit is meant to specialize on understanding the specific style and semantics of the twitter domain, providing a specialized model with plenty of possibilities, as it is the case of disinformation analysis. Thus, as disinformation spreads on OSNs, it is extremely relevant to understand the language used in this modality of communication, which usually differs from actual news articles, encyclopedic pages or scientific journals. The following pre-training is designed to bridge the issues identified in Section \ref{sec.sota}, namely specialization in a language and specialization on twitter text. 

An overview of the steps taken to pre-train and fine-tune BERTuit are summarized on Figure \ref{fig.pipe}. In short, this is a typical masked language modeling (MLM) pre-train pipeline tuned with several adjustments to succeed in this domain. We detail all elements in the pipeline in order of appearance:

\begin{figure}[h]
    \centering
    \includegraphics[width=0.5\textwidth]{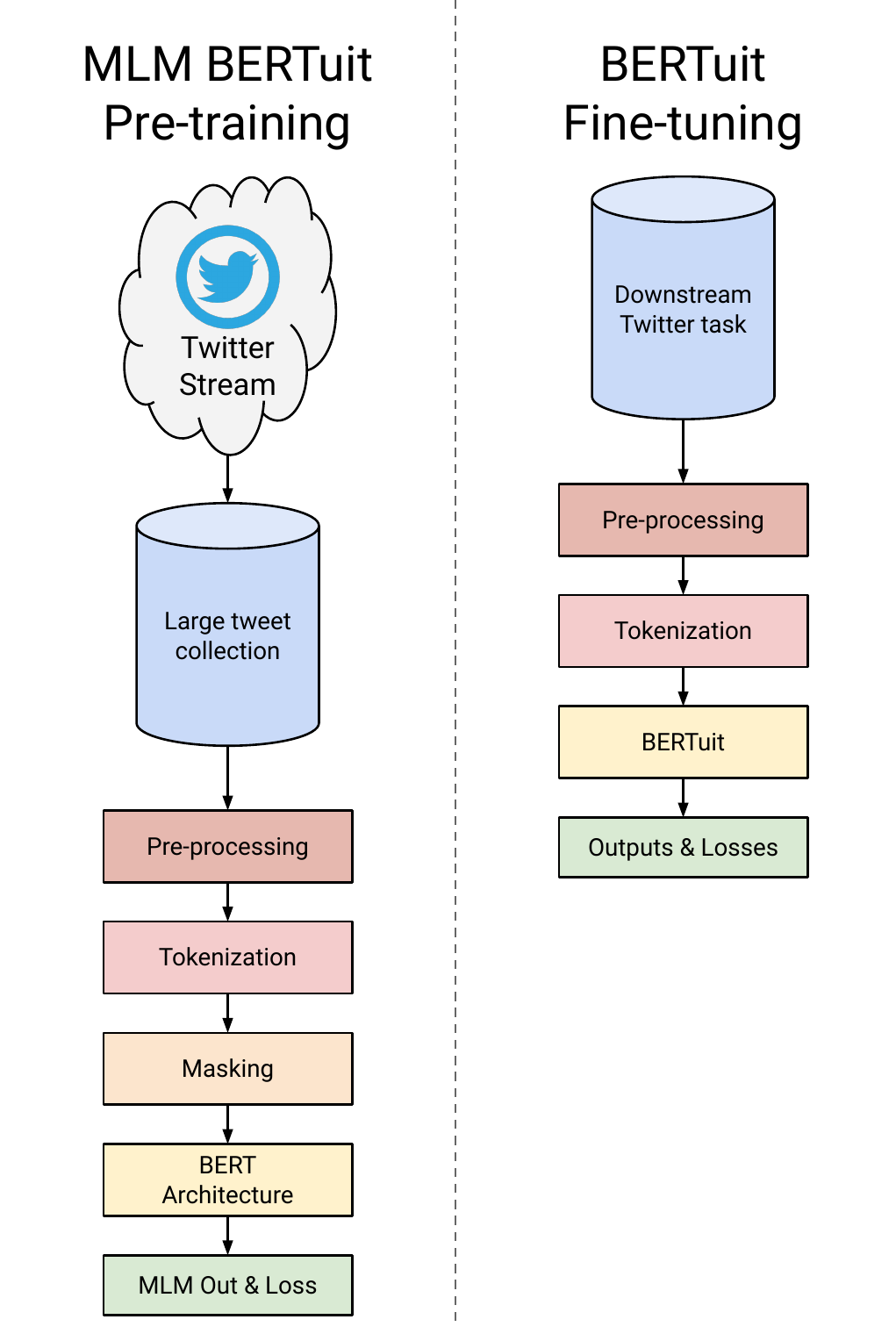}\caption{\textit{Left}: BERTuit pipeline for Masked Language Modeling objective. \textit{Right}: BERTuit pipeline to fine-tune on downstream tasks compatible with Spanish Twitter data.}
    \label{fig.pipe}
\end{figure}

\begin{enumerate}
    \item \textbf{Twitter Stream Grab~\cite{BibEntry2022Jun}:} The Archive \footnote{Archive Twitter Stream Grab address \url{https://archive.org/details/twitterstream}} Twitter stream Grab is our main data source. It contains twitter information from authors, text and, most importantly, language used. Their data has been archived since 2012, using shallow information from the social network. This means that most of the information in this large corpora is composed of top-level tweets, excluding responses or citations. This is one of the few limitations of the corpus, however as there is a massive amount of data from several hundreds of accounts, it is a minor issue.
    
    \item \textbf{Large Tweet Collection:} Using the specified data gathering we have selected a recent sample from May 2018 to January 2021. The only criteria to select data from this sub-sample has been \textit{1)} As our goal is to produce a natively trained language model, only tweets marked as Spanish language are extracted; and \textit{2)} exclude url-only tweets, as Twitter shortens urls thus stripping any possible meaning from the text. The result of this selection is a corpus with >$230$ million tweets.
    
    \item \textbf{Pre-processing:} Two minor alterations have been made on the corpus. Both urls (anything beginning with \textit{https} or \textit{http}) and user tags (string of thext with shape \textit{@user}) are substituted by special sequences called \textit{<usr>} and \textit{<url>}. Our reasoning behind these replacements is that either users or urls lack any relevant semantic charge beyond their positioning in the text, as users can change their names at any time, and urls are shortened to be illegible.
    
    No more alterations are made to the text, including the sequence length. We acknowledge that emoji and mistypings are a fundamental part of online communications, and as such, contain semantic meaning. Transforming emoji to text would destroy some of the meaning, as they are employed to convey complex feelings or expressions. Mistypings on the other hand are common as many user prefer to use shorthand writing, or are not educated enough to properly write some words; these preferences and errors usually convey significance to the text and as such no correcting effort has been made.
    
    \item \textbf{Tokenization:} This process transforms text to recognizable indices by the lexical embedding layer of the transformer architecture. As twitter language is extremely mutable and prone to error we develop a Byte Pair Encoding (BPE) dictionary comprised of $30$ thousand tokens. Furthermore, the special sequences \textit{<usr>} and \textit{<url>} are encoded with their own special token.
    
    \item \textbf{Masking:} In short, self-supervised training consists on learning to detect information that has been stripped from the original input. For Masked Language Modeling, a random set of tokens are replaced by a special \textit{<mask>} token, which the language model is meant to rebuild. This process in practice this trains the model to infer words from their neighboring context, which in turn results in robust representations of input tokens. It is also notable that masked tokens with a model trained this way can produce a set of probabilities for the most likely words to appear in the mask gap; in our case this is not limited to words, but emoji is allowed to be masked and predicted.
    
    On RoBERTa, it was shown that the Next Sentence Prediction objective did not significantly impact results for later fine-tuning on downstream tasks, therefore relying on Masked Language Modeling (MLM) is enough to pre-train. Following the original BERT paper, we select 15\% of the non-special tokens to be candidates for masking. From the candidates, 80\% are replaced by masks, 10\% are replaced by another random token and the remainder are left unchanged. The masked tokens are used as labels to perform the pre-training, while the replaced and unchanged tokens are not used in the output.
    
    To build batches of masked sequences we build padded sequences of length $256$ tokens. No individual tweet ever surpasses this length as they are limited to 260 characters, on average resulting on $100$ tokens for the sequence length. We are interested in the model learning individual Tweets from specific users, as disinformation is disseminated by both people and bots. Tweets have to be processed individually, because unrelated information from a tweet could pollute the word embeddings of another tweet if they are present in the same sequence. While this is not a problem for generalist models such as RoBERTa or GPT, we deem it counter-productive to our model as our focus is on specialization in this particular domain, despite being more computationally efficient than our alternative. 
    
    \item \textbf{BERT Architecture:} Again, and following RoBERTa best practice indications, we use the default base BERT architecture with minor modifications. Following the previous sequence length choice, we limit the size of the static positional embeddings of BERT to $256$. We leave this additional large sequence length compared to the domain to perform sentence pair tasks such as stance detection or language inference, where two twits could be concatenated to perform classification. Hyperparameters of the topology are given in Table \ref{tab.hyperparameters-topology}, hidden and feedforward sizes are $768$ and $3072$, with $12$ heads and $12$ transformer blocks, regularized by a dropout of $10\%$.
    
\begin{table}[]
\centering
\begin{tabular}{@{}rl@{}}
\toprule
\textbf{Hyperparameter}        & \textbf{Value} \\ \midrule
\textit{Hidden layer size}     & $768$            \\
\textit{Feedforward size}      & $3072$           \\
\textit{Positional embeddings} & $256$            \\
\textit{Attention heads}       & $12$             \\
\textit{Number of blocks}      & $12$             \\
\textit{Vocabulary}            & $30000$          \\
\textit{Global dropout (\%)}   & $10\%$           \\ \bottomrule
\end{tabular}\label{tab.hyperparameters-topology}
\caption{Hyperparameters chosen for the topology, adapted from BERT}
\end{table}

    \item \textbf{MLM output, loss and optimization:} The architecture produces a set of words from the input twit sequence, which correspond to each token. Only \textit{<mask>} tokens are considered to learn, as such the average cross-entropy loss for each masked token is computed. The optimization process, shown in Table \ref{tab.hyperparameters-topology} follows the steps of RoBERTa and BERT: the pre-training model is run for $1e6$ training steps, the learning rate is scheduled with a warm-up period of $1e4$ steps and a later decay, where the learning rate peaks at $1e$-$4$. The batch size is established as $256$ by using gradient accumulation. Adam is used as the optimizer with the reported scheduling and momentum terms set to $\beta_1=0.9$ and $\beta_2=0.999$, without weight decay. The model takes approximately 2 weeks to train on a single GPU.
    
\begin{table}[]
\centering
\begin{tabular}{@{}ll@{}}
\toprule
\multicolumn{1}{r}{\textbf{Hyperparameter}} & \textbf{Value}               \\ \midrule
\multicolumn{1}{r}{\textit{Training steps}} & $1e6$                          \\
\multicolumn{1}{r}{\textit{Optimizer}}      & Adam                         \\
                                            & $\alpha_{peak}=1e$-$4$         \\
                                            & $\beta_1=0.9$                \\
                                            & $\beta_2=0.999$              \\
\multicolumn{1}{r}{\textit{Scheduler}}      & Linear decay with warmup \\
                                            & Warmup steps $= 1e4$           \\
                                            & min $\alpha = 0$        \\
\multicolumn{1}{r}{\textit{Batch size}}     & $256$                          \\ \bottomrule
\end{tabular}\label{tab.hyperparameters-optimization}
\caption{Hyperparameters chosen for optimization, adapted from BERT and RoBERTa}
\end{table}
     
     \item \textbf{Inference:} Finally, after the pre-training is successful, BERTuit can be run for fine-tuning and inference. To run the model for either mode, the preprocessing and tokenization step must be done. In the case of fine-tuning, an additional head must be added and trained to perform other tasks. While BERTuit is very familiar with the Twitter language, it has to be fine-tuned to perform adequately in downstream tasks. The main advantage of this model is that it can generalize on very low amounts of fine-tuning data, due to the very close proximity of the domain language (spanish and twitter expression).
\end{enumerate}

\begin{table}[]
\centering
\resizebox{1\textwidth}{!}{%
\begin{tabular}{@{}ccl@{}}
\toprule
\textbf{Identifier} & \textbf{N} & \multicolumn{1}{c}{\textbf{Hoax}} \\ \midrule
0 & 132 & Messenger RNA vaccines can make us transgenic \\
1 & 250 & COVID-19 vaccines cause seizures \\
2 & 548 & The United States admitted that only 6\% of reported deaths were actually from coronavirus \\
3 & 149 & Face masks cause neurodegenerative diseases \\
4 & 103 & An image of a patent in the Netherlands for a method to "test COVID-19" since 2015 \\
5 & 963 & The coronavirus vaccine can leave you sterile \\
6 & 152 & There is a plan designed for COVID-19 since 2017 in World Bank documents \\
7 & 221 & The COVID-19 vaccine has been found to permanently destroy our immune system \\
8 & 241 & Drinking lots of water and gargling with hot water and salt eliminates the coronavirus \\
9 & 272 & It is recommended to keep the body in an alkaline state \\
10 & 114 & Eucalyptus prevents or eliminates the new coronavirus \\
11 & 218 & Guava tree leaf may prevent or reverse effects of COVID-19 \\
12 & 73 & NASA listed chlorine dioxide as a universal antidote in 1988 \\
13 & 138 & Drinking wine can be beneficial against COVID-19 \\
14 & 113 & The use of the mask causes deaths from bacterial pneumonia \\
15 & 99 & Vitamin C prevents the virus \\
16 & 579 & Christine Lagarde said: The elderly live too long and that is a risk to the global economy \\
17 & 211 & \begin{tabular}[c]{@{}l@{}}There is a relationship between the Chinese biological laboratory in Wuhan, the pharmaceutical \\ companies Glaxo and Pfizer and people like George Soros and Bill Gates among others\end{tabular} \\
18 & 368 & The coronavirus dies at 27º \\
19 & 127 & Scientist Charles Libier was arrested for creating the Covid-19 coronavirus. \\ \bottomrule
\end{tabular}%
}
\caption{Selection of 20 hoaxes for the visualization. They were extracted in Spanish and, for purposes of this article, translated into English. Identifiers coincide with the crosses of the visualization. N is the number of tweets related to the hoax}
\label{tab.hoaxes}
\end{table}

\begin{table}[]
\resizebox{\textwidth}{!}{%
\begin{tabular}{@{}rlllll@{}}
\toprule
\textbf{Task}                & \textbf{Metric}                        & \textbf{BERTuit}         & \textbf{M-BERT} & \textbf{XLM-RoBERTa} & \textbf{XLM-Twitter}    \\ \midrule \midrule
\multirow{2}{*}{Hate speech detection}          & \multicolumn{1}{l|}{\textit{Accuracy}} & \textbf{0.8275 (0.011)}  & 0.7864 (0.012) & 0.7825 (0.017)  & 0.8115 (0.015)  \\
                             & \multicolumn{1}{l|}{\textit{F1-Score}} & \textbf{0.7728 (0.01)}   & 0.7022 (0.018)  & 0.6852 (0.027)       & 0.7553 (0.017)          \\ \midrule
\multirow{2}{*}{Irony detection}                & \multicolumn{1}{l|}{\textit{Accuracy}} & \textbf{0.7431 (0.0083)} & 0.7037 (0.007) & 0.7297 (0.0095) & 0.7421 (0.0078) \\
                             & \multicolumn{1}{l|}{\textit{F1-Score}} & \textbf{0.7429 (0.0083)} & 0.7035 (0.0074) & 0.7295 (0.0096)      & 0.741 (0.008)           \\ \midrule
\multirow{2}{*}{Issue detection}                & \multicolumn{1}{l|}{\textit{Accuracy}} & \textbf{0.6828 (0.031)}  & 0.5604 (0.061) & 0.549 (0.078)   & 0.5984 (0.038)  \\
                             & \multicolumn{1}{l|}{\textit{F1-Score}} & \textbf{0.6473 (0.032)}  & 0.518 (0.052)   & 0.4632 (0.11)        & 0.531 (0.039)           \\ \midrule
\multirow{2}{*}{NER}         & \multicolumn{1}{l|}{\textit{Accuracy}} & \textbf{0.9622 (0.0036)} & 0.9519 (0.0065) & 0.9516 (0.0023)      & 0.9604 (0.003)          \\
                             & \multicolumn{1}{l|}{\textit{F1-Score}} & \textbf{0.5634 (0.022)}  & 0.5178 (0.021)  & 0.5262 (0.015)       & 0.5595 (0.017)          \\ \midrule
\multirow{2}{*}{POS tagging} & \multicolumn{1}{l|}{\textit{Accuracy}} & 0.8518 (0.14)            & 0.8448 (0.13)   & 0.8484 (0.14)        & \textbf{0.8554 (0.14)}  \\
                             & \multicolumn{1}{l|}{\textit{F1-Score}} & 0.8062 (0.084)           & 0.7901 (0.084)  & 0.8039 (0.084)       & \textbf{0.8105 (0.085)} \\ \midrule
\multirow{2}{*}{Sentiment analysis}             & \multicolumn{1}{l|}{\textit{Accuracy}} & \textbf{0.6585 (0.0058)} & 0.6521 (0.007) & 0.6107 (0.025)  & 0.6281 (0.049)  \\
                             & \multicolumn{1}{l|}{\textit{F1-Score}} & \textbf{0.6325 (0.0071)} & 0.6248 (0.0075) & 0.5772 (0.027)       & 0.5987 (0.054)          \\ \midrule
\multirow{2}{*}{Token-level sentiment analysis} & \multicolumn{1}{l|}{\textit{Accuracy}} & \textbf{0.8885 (0.016)}  & 0.7139 (0.26)  & 0.6202 (0.4)    & 0.8833 (0.0087) \\
                             & \multicolumn{1}{l|}{\textit{F1-Score}} & 0.4785 (0.018)           & 0.3527 (0.13)   & 0.2805 (0.17)        & \textbf{0.4788 (0.013)} \\ \bottomrule
\end{tabular}%
}
\caption{Summary of results for accuracy and f1-score, with the average and deviation in parentheses. Best results for each row are marked with bold}
\label{tab:0-result_summary}
\end{table}

\section{BERTuit Validation} \label{sec.validation}
Before delving into possible applications of the BERTuit model, we aim to empirically explain its performance. To this end we run BERTuit on some common tasks for Twitter analysis. We measure a series of quality metrics appropriate to classification problems. We use other state-of-the-art models fit to this task and a comparison against the best reported results (if available).

\subsection{Task description} \label{ssec.tasks}
We find seven tasks to evaluate: two binary sequence classification, two for multi-class sequence classification and three for multi-class token classification. The following tasks have been performed. By default, if no validation or test is specified, the data is split into a $70/10/20$ proportion for training, validation and testing. Unless noted, this is the default pick to evaluate each model.

\begin{enumerate}
    \item \textbf{Hate Speech detection}: This task consists of finding traces of hateful content in a sequence, as such it is a binary sequence classification task. The dataset used is HaterNet, containing >$6$ thousand tweets~\cite{lara_quijano_sanchez_2019_2592149} labelled where hate speech is present. 
    \item \textbf{Irony detection}: Finding humor is a common task when recognizing text, which is more prevalent on OSNs. Irony detection is a binary sequence classification task, where the model has to find whether a text has been written ironically or sincerely. To test this task we use >$14$ thousand tweets annotated with irony~\cite{Mezaruiz2017}.
    \item \textbf{Issue detection}: This problem consists on multi-class sentence classification, labeling examples depending on the overall coarse-grain topic mentioned. The dataset used consists of >$3$ thousand tweets from the 2015 Spanish General Election~\cite{baviera2019twitter}, differentiating between several issues: Political, Policy, Campaign, Personal or Other. The labeling is directly related to the election and the topics are intertwined. 
    \item \textbf{Named entity recognition}: Named entity recognition (NER) consists on finding items within the text that refer to named places, people, among other labels. This classification is performed at token level, requiring finer granularity than the previous tasks. In this case we use the xLiMe~\cite{reixlime2016} which contains >$300$ thousand tokens from >$20$ thousand texts, annotated for three relevant tasks: NER, part-of-speech and token-level sentiment analysis. For NER, nine labels are found: people, location, organization, miscellaneous and nothing; where the first four labels can be the beginning of such label or a continuation, summing up to nine entity types.
    \item \textbf{Part-of-speech tagging}: Part-of-speech (POS) consists on labeling tokens with their syntactical tag, for example: verbs, nouns and so on. Using xLiMe we find several possible tags: verb, noun, adverb, adjective, pronoun, adposition, determiner, punctuation, user mentions, urls, continuation numbers and emoticons.
    \item \textbf{Sentiment Analysis}: Multi-class sentence classification consisting on finding the sentiment of a twit, either neutral, negative or positive. We use a public dataset of >$270$ thousand tweets~\cite{mozetivc2016multilingual, mozetivc2018evaluate}, this dataset contains a multi-lingual modality, however we perform experiments on the monolingual task. No test set is provided, therefore the data is split.
    \item \textbf{Token-level Sentiment Analysis}: This task is very similar to sentiment analysis however, this time it is performed token by token. As a multi-class token classification each token has been labeled with an emotional polarity of positive, neutral or negative; which we can also find on the xLiMe corpus. 
\end{enumerate}

\subsection{Baseline fine-tuning} \label{ssec.base}
To perform a fair comparison against the state of the art methods, we perform experiments against the following models. They have been selected because of their multilinguality and, in the case of XLM-Twitter, because it has been built specifically for Twitter domain texts, alongside its multilingual capabilities.
\begin{enumerate}
    \item \textbf{M-BERT}: Multilingual BERT stems from the original BERT paper, with the ability to recognize 100 languages, Spanish among them. This is achieved using BERT pre-training objectives and switching the pre-training corpus to a multi-lingual set of texts.
    \item \textbf{XLM-RoBERTa}: Originally XLM~\cite{lample2019cross} was designed for multi-language pre-training, however, as it happened with BERT and RoBERTa, it was found that a more robust optimization process was possible and it was used on XLM, leading to XLM-RoBERTa. As with M-BERT it is able to recognize 100 languages.
    \item \textbf{XLM-Twitter}: BERTweet is a model trained with twitter data exclusively, however it is unable to recognize multiple languages. Using XLM-RoBERTa optimization with 100 languages on exclusively twitter data, it is possible to bridge BERTweet shortcomings \cite{barbieri2021xlm}.
\end{enumerate}

All models described are fine-tuned under the same conditions. For Sequence classification tasks Adam optimization is used with a learning rate of $2e$-$5$ during $3$ epochs. For token classification added time and rate is needed, with a learning rate of $5e$-$5$ and $10$ epochs. The learning rate is linearly decayed for the duration of the fine-tuning. Early stopping is performed if $5$ epochs pass without improvements in validation, however this rarely happens. For all tasks labels are balanced through class weights, to account for any class imbalance within each dataset.

To measure the quality of the models we measure both accuracy and f1-score, as some problems have unbalanced classes we prioritize macro-average f1-score. Either are reported as pairs of average and deviation, as we perform 10 runs for each model and task, meant to reduce the influence of classification weights random initialization.

\subsection{Experimental results} \label{ssec.results}

The results from the described methodology are presented on table \ref{tab:0-result_summary}. To begin with, BERTuit outperforms every other model at most scenarios except POS tagging and token-level sentiment analysis. Analysing the runner-up on both tasks we observe that deviations (F1-score and accuracy) of BERTuit and XLM-Twitter heavily overlap, pointing at large variations of the quality of predictions across runs which is specially pointed on POS tagging. Other domains show very pointed improvements, such as the Issue detection task, where all other models are outperformed by 10 points by BERTuit. Others like hate speech are outperformed by 2 points or sentiment analysis by at least 1.

The most pointed result from our experimentation is that, even some models drop in performance for some domains while improving on others, the usage of BERTuit remains consistent across all tasks, improving greatly upon the best available baseline or tying against it in the worst case scenario. It is also worth noting that on some tasks where results are unstable for other models, BERTuit offers more consistent results, this is the case for token-level sentiment analysis or issue detection, offering added stability over the alternatives. To summarize, our approach offers improved performance against state-of-the-art baselines and has a more consistent fine tuning process.

To further argue the strengths of BERTuit, the time taken to fine-tune each model has also been measured across runs. On table  \ref{tab:1-time_summary} the average time in seconds is shown, it is observed that, for sequence classification tasks, BERTuit is faster than the baselines. On the other hand, for sequence classification tasks BERTuit is slower. The added pre-processing of the tokenizer and further inefficiencies with labelling could explain this loss on performance.

\begin{table}[]
\resizebox{0.9\textwidth}{!}{%
\begin{tabular}{@{}rllll@{}}
\toprule
\textbf{Task}                        & \textbf{BERTuit}      & \textbf{M-BERT}      & \textbf{XLM-RoBERTa} & \textbf{XLM-Twitter} \\ \midrule \midrule
\multicolumn{1}{r|}{Hate speech detection} & \textbf{23.75 (12.6)}   & 28.95 (12.5)   & 30.50 (12.4)   & 30.22 (12.6)   \\
\multicolumn{1}{r|}{Irony detection} & \textbf{36.72 (12.9)} & 46.84 (12.5)         & 50.28 (12.5)         & 49.43 (12.8)         \\
\multicolumn{1}{r|}{Issue detection} & \textbf{15.39 (12.7)} & 18.65 (12.8)         & 19.34 (12.7)         & 19.28 (12.4)         \\
\multicolumn{1}{r|}{NER}             & 46.76 (7.8)           & \textbf{32.68 (8.0)} & 37.86 (7.9)          & 37.86 (7.9)          \\
\multicolumn{1}{r|}{POS tagging}     & 47.34 (8.4)           & \textbf{33.20 (8.5)} & 38.36 (8.1)          & 38.28 (8.3)          \\
\multicolumn{1}{r|}{Sentiment analysis}    & \textbf{1016.57 (16.5)} & 1507.16 (13.8) & 1554.46 (17.6) & 1552.78 (16.0) \\
\multicolumn{1}{r|}{Token-level sentiment analysis}& 46.52 (8.3)           & \textbf{32.97 (8.7)} & 38.24 (9.0)          & 37.68 (8.1)          \\ \bottomrule
\end{tabular}%
}
\caption{Summary of average training time required to fine-tune a model to a task in accordance to the described hyper-parameters. The lowest values are marked in bold; averages are presented along their respective standard deviation.}
\label{tab:1-time_summary}
\end{table}

\begin{figure}[htpb]
    \centering
    \includegraphics[width=0.9\textwidth]{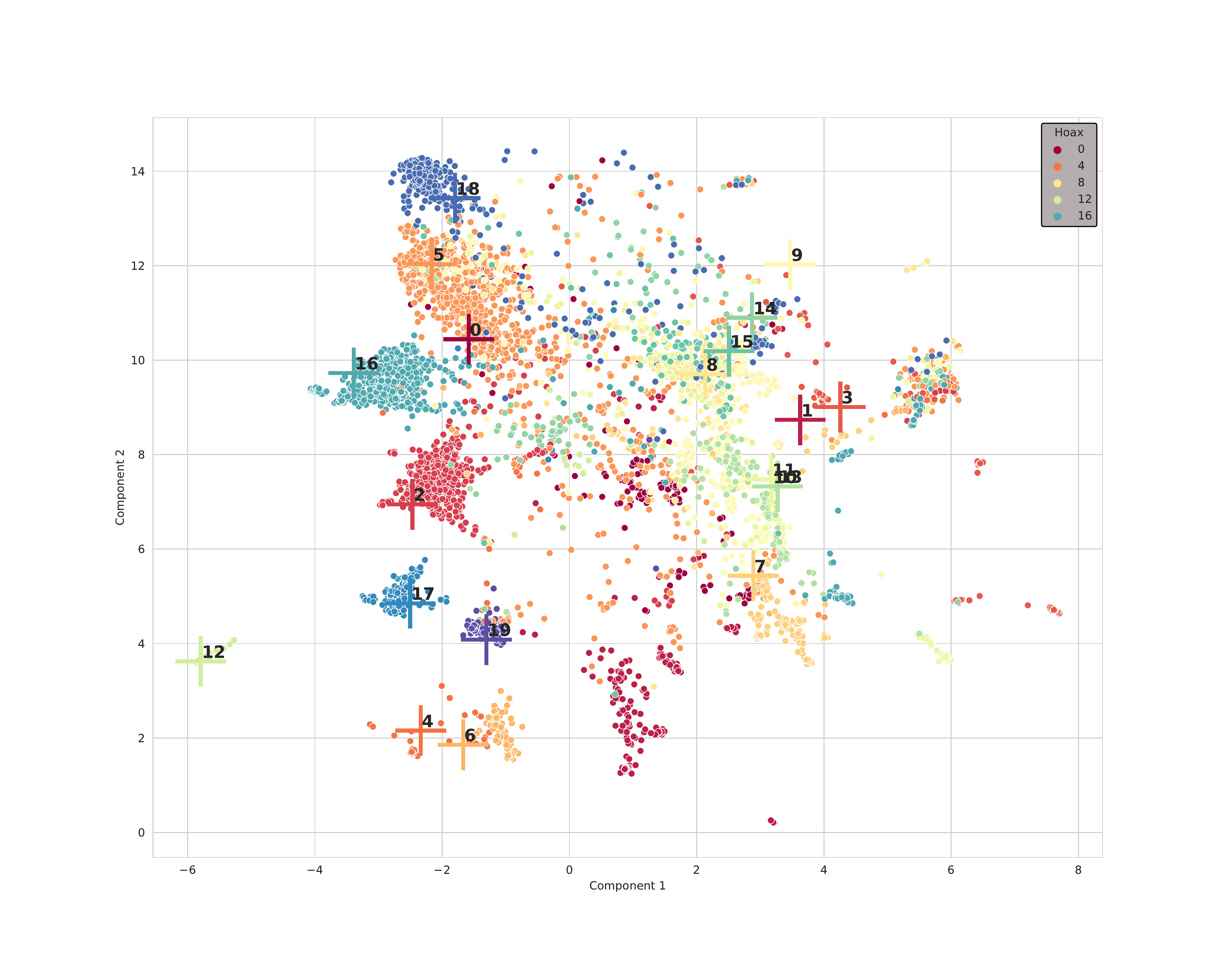}\caption{Projection of hoaxes, legend represent the hoax number, also marked as crosses over the scatterplot. Axes x and y represent the projection value generated by UMAP from the original embeddings. Points that share hue with a cross are claims (point) related to a hoax (cross) with the same hue.}
    \label{fig.repr}
\end{figure}

\section{Applications} \label{sec.apps}
Through the use of BERTuit we are able to better understand Twitter, as it is shown on Section \ref{ssec.results}. With this novel tool, applications for misinformation detection, tracking and countering can be developed. Our analysis focuses mainly on text to showcase BERTuit and its utility, multimodal architectures are possible and encouraged to build stronger tools. 

\subsection{Visualizing related misinformation} \label{ssec.simil}
Misinformation texts share certain stylistic choices that makes determining the topic of a false statement difficult. A strong enough model should be able to differentiate across topics of misinformation. We aim to provide a simple method to characterize visually the topic of misinformation tweets. We collected >$12$ thousand tweets supporting 61 COVID-19 popular hoaxes. Each claim is embedded using BERTuit, extracting the second to last hidden state of the transformer. To obtain a sentence embedding we extract the average pooling of all non-padding tokens.

Projecting embeddings onto a 2D space for visualization can be performed via dimensionality reduction techniques, such as Uniform Manifold Approximation and Projection (UMAP)~\cite{mcinnes2018umap}. The parametrization used for this application consists of 2 components, 500 neighbors, a minimal distance of $0.25$ and the cosine distance as the metric. A selection of 20 hoaxes and $5$ thousand associated claims are selected for a cleaner visualization in a 2D space. The result of projecting hoaxes and tweets is presented on Fig. \ref{fig.repr}, and the translated hoaxes referenced in the graph are presented on Table \ref{tab.hoaxes}.

Some claims in the visualization are neatly grouped, with the embedding of the hoax among them. For example, hoaxes 2, 12 and 17 form their own groups with very little overlap with other hoaxes. Others like hoax 5, 16 and 18 end up closer together but with clearly defined borders. We find an interesting case study on hoaxes 10, 11 and 13 which share the same space on the map because all of them relate to food-based ineffective remedies such as guava, eucalyptus and wine. Hoax number 9 is another interesting case, though the embedding of the hoax itself is far from where it should be, its associated tweets are grouped with other similar diet-based remedies such as hoaxes 10, 11 and 13, sharing some area also with hoax 8. This method is not perfect, as some hoaxes are not well positioned, such as hoax 0, 1 or 9, but their related hoaxes maintain either some separation from other groups (such as tweets related to 0) or are grouped with very related topics (such as tweets related to hoax 8).

\subsection{Profiling fake news spreaders} \label{ssec.authors}

\begin{table}[htpb]
\centering
\begin{tabular}{@{}lccc@{}}
\toprule
 & \multicolumn{1}{l}{\textbf{Accuracy}} & \multicolumn{1}{l}{\textbf{Precision}} & \multicolumn{1}{l}{\textbf{Recall}} \\ \midrule
\textbf{Reduce mean}        & \textbf{81.90\%} & 80.15\% & 79.10\% \\
\textbf{Reduce max }        & 75.30\%          & 73.36\% & 75.50\% \\
\textbf{Bi-LSTM }           & 79.40\%          & 79.40\% & 79.40\% \\
\textbf{PAN 20 - Top score} & 80.50\%          & -       & -       \\ \bottomrule
\end{tabular}
\caption{Summary of the results obtained in the PAN 20 competition on Profiling Fake News Spreaders on Twitter 2020, showing the top result in the competition and the three strategies proposed based on embeddings generated with BERTuit
}
\label{tab:spreaders}
\end{table}

\begin{figure}[htpb]
    \centering
    \includegraphics[width=1\textwidth]{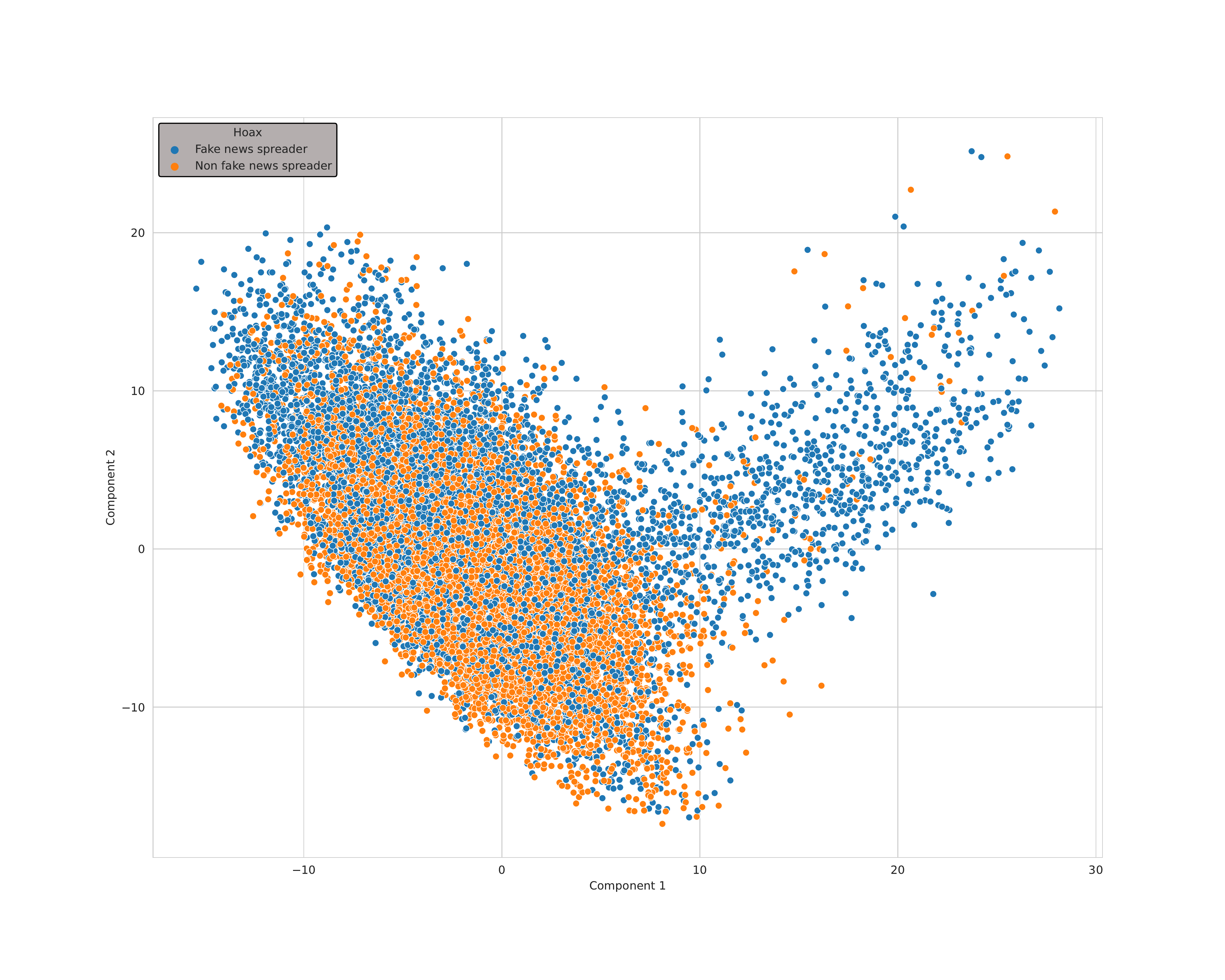}
    \caption{Projection of the embeddings generated with BERTuit for all the tweets in the test set of the Profiling Fake News Spreaders on Twitter 2020 competition.}
    \label{fig:tweets_projection}
\end{figure}

\begin{figure}[htpb]
    \centering
    \includegraphics[width=1\textwidth]{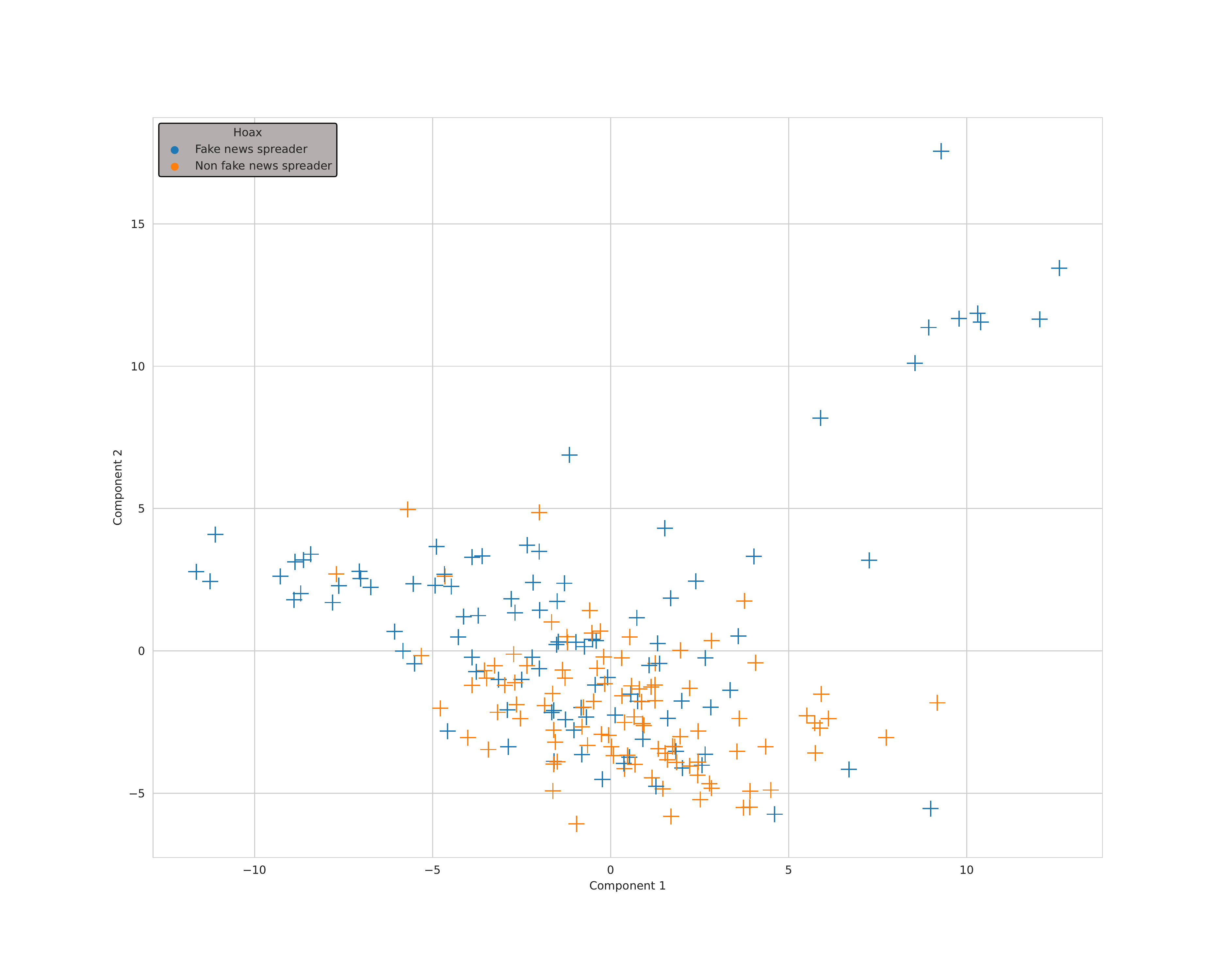}
    \caption{Projection of the embedding generated for each author as the mean vector of the embeddings obtained with BERTuit for all the tweets in the test set of the Profiling Fake News Spreaders on Twitter 2020 competition.}
    \label{fig:authors_projection}
\end{figure}

This second application focuses on a highly relevant issue in the disinformation domain, the profiling of fake news spreaders. At CLEF 2020, a competition was proposed in this line called Profiling Fake News Spreaders on Twitter 2020~\cite{rangel2020profiling}. The competition provided a dataset of Twitter users labelled according to if they spread or not false information. The dataset includes 300 users for training, each one with 100 tweets. Although the dataset also included an English set of data, we only evaluate on the Spanish part due to the objectives of BERTuit.

Before training a classification head on top of the BERTuit architecture to distinguish between fake news spreaders and non-spreaders, it is necessary to adopt an strategy to obtain a representation of the whole author based on his/her tweets. While we can use BERTuit to obtain a representative word embedding vector of each tweet, in order to combine all the tweets and represent the whole authors we identified three possibilities:

\begin{itemize}
\item \textbf{Average of tweets embeddings:} An average of all the embeddings of the tweets of the author can produce a new single embedding with a full representation of the author that can be later used to identified if it is a fake news spreader.
\item \textbf{Maximum of tweets embeddings:} The maximum of elements across dimensions can also be used to reduce the set of tweets embeddings to one representative author embedding.
\item \textbf{Sequence of embeddings:} Instead of reducing the tweets embeddings to one representative embedding for the whole author, it is also possible to use them as a sequence of embeddings, thus training a recurrent architecture to distinguish between sequences.
\end{itemize}

We defined two different architectures to be trained on these representations:

\begin{itemize}
\item \textbf{A dense architecture: } composed of two hidden layers with 60 neurons an hyperbolic tangent activation function to be trained with a unique representation vector by author (using the average or the maximum representation strategy).
\item \textbf{An LSTM architecture:} with two bidirectional LSTM layers of 80 neurons and a batch normalization layer, to be trained with the third representation of the author (sequences of embeddings).
\end{itemize}

The averaged results of 10 executions are displayed in Table~\ref{tab:spreaders}. As can be seen, the average of the embeddings of each user's tweet provides the most appropriate representation, reaching a 81,90\% accuracy, which allows to improve the previous top score in the competition. The use of Bi-LSTM architecture produces a lower result, meaning that this architecture is not able to extract sequential patterns from the data.

To better understand how the embeddings of the tweets collect relevant characteristics, we follow the same procedure used in previous section to build a 2D projection, in this case through a PCA, given that it shows a better distribution in the space. As can be seen in Fig.~\ref{fig:tweets_projection}, the projection of the tweets presents a large overlapping in the range between -10 and 10 for both components, while a section of tweets from fake news spreaders separates from this region in right part of the plot. This behaviour can be attributed to the complex language used in Twitter, and that many tweets present strong similarities between authors of both classes. 

To evaluate the average embedding generated for each author, a projection of these vectors is displayed in Fig.~\ref{fig:authors_projection}. As in the case of the projection of the tweets embeddings, the distribution of the authors follows a similar pattern. Most of the authors posting real content are placed in the area between -5 and 10 in the horizontal axis and -5 and 0 in the vertical one. Despite several misinformation spreaders placed in this area, most of them are located elsewhere. Nevertheless, both previous figures evidence the complexity of this scenario, where it is possible to find authors and tweets of both labels located very close in the representation space.

\begin{figure}[htpb]
    \centering
    \includegraphics[width=0.8\textwidth]{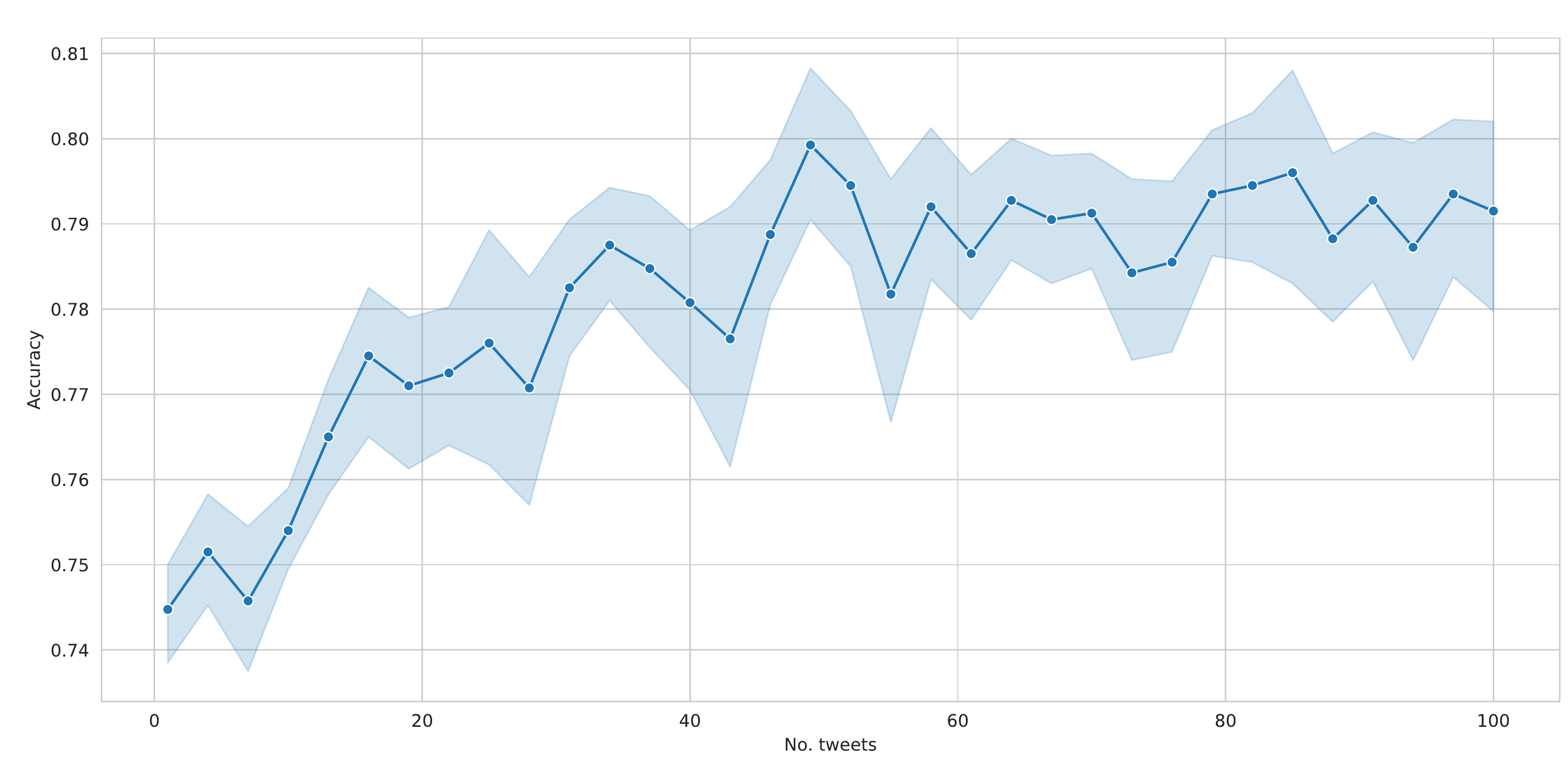}
    \caption{Evaluation of the performance of BERTuit in the detection of Fake News Spreaders on Twitter 2020 PAN competition according to the number of tweets considered for each author.}
    \label{fig:accuracy_number_tweets}
\end{figure}

\begin{figure}[htpb]
    \centering
    \includegraphics[width=1\textwidth]{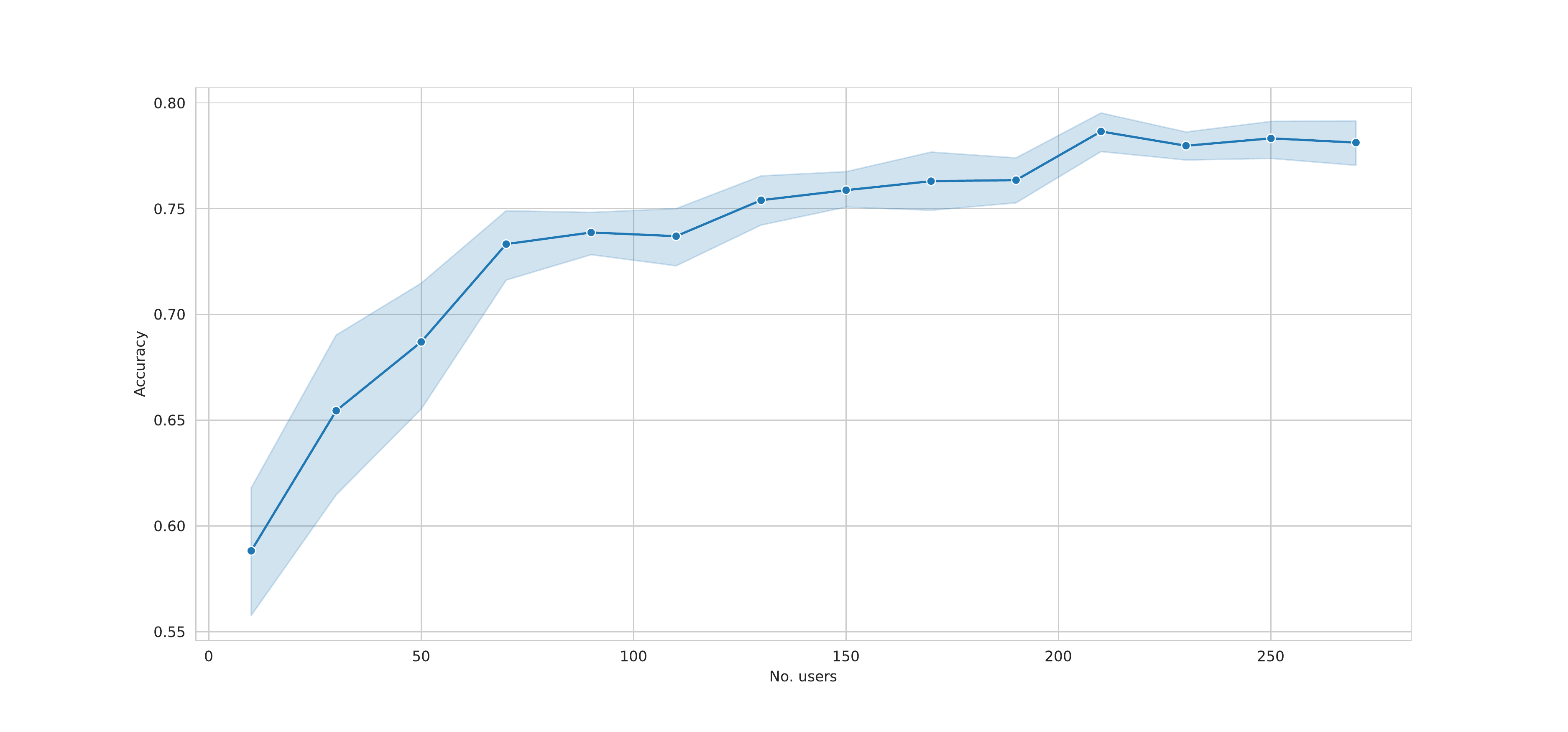}
    \caption{Evaluation of the performance of BERTuit in the detection of Fake News Spreaders on Twitter 2020 PAN competition according to the number of users considered during the training step.}
    \label{fig:accuracy_number_users}
\end{figure}

Although the two-dimensional projection presented allows to derive important details, it does not enable to understand the important characteristics that a classification model chooses in this domain, which be related to the semantics, the style or the use of specific words, among other uses. To better understand which are the most helpful features in a classification process, we performed two experimenta, aimed at assessing the individual contribution of each tweet to the labelling of the whole author and the information collected from different authors required to understand the difference between both labels. The first evaluation aims to identify the contribution of information considered from each user. The results, displayed in Fig.~\ref{fig:accuracy_number_tweets}, show the evolution of the accuracy according to the number of tweets of each author used during the training phase. Although with a low number of tweets the model shows lower rates of accuracy, collecting just one tweet provides a fruitful information source to label that user, reaching more than 74\% accuracy. It can also be seen that a higher number of tweets does not involve higher accuracy rates. These two facts evidence the ability of BERTuit to extract relevant patterns from small sets of data.

Additionally, we also evaluated the impact of the number of users in the classification performance. Fig.~\ref{fig:accuracy_number_users} shows the evolution of the accuracy according to the number of users considered during the training phase. As expected, there is a strong correlation between both values. In contrast to the information from each user (the number of tweets), the number of different authors plays an essential role in the training process of a classification model. Thus, it must be prioritized to collect a varied set of data instead of extracting large amounts of data from individual users.

\section{Conclusions and future Work} \label{sec.conclusions}
In this work we have described the limitation of current language models to address tasks in specific scenarios where language adopts a very specific form, exhibiting characteristics that are not present in other information sources. In addition, the use of multilingual models also presents problems in this domain. In order to address all these issues, in this paper we have presented BERTuit, a language model trained with a a RoBERTa optimization in a corpus composed of 230M tweets in Spanish. Our goal is to provide a useful instrument for those interested in developing Natural Language Processing solutions where the input language is generated in a OSN, thus dealing with leetspeak, abbreviations or specific words that are specific of these platforms. To validate BERTuit, we performed a thorough evaluation against state-of-the-art alternatives in tasks highly related to Social Networks, such as Hate Speech detection, Irony detection or Sentiment Analysis, reliably obtaining better results on most downstream tasks tested. In a second step, we have described in detail two different applications for BERTuit, one of them focused on visualizing related misinformation, showing how our model can be used to generated powerful representations of tweets through embedding vectors in combination with a dimensionality reduction technique, providing a two-dimensional projection that enables to obtain useful conclusions from the data. Another application is the profiling of fake news spreaders on Twitter, showing the ability of the model to detect this type of user by generating a representation of each author by averaging the embedding vector of each tweet of that user. In future work, our goal is to extend the concept of a specialised language model in OSNs to a multilingual scenario, assessing the ability of these models to understand different languages in this complex domain simultaneously, and also further exploring the applications of BERTuit.

\section{Acknowledgements} \label{sec.ack}
This work has been supported by the research project CIVIC: Intelligent characterisation of the veracity of the information related to COVID-19, granted by BBVA FOUNDATION GRANTS FOR SCIENTIFIC RESEARCH TEAMS SARS-CoV-2 and COVID-19, by the Spanish Ministry of Science and Innovation under FightDIS (PID2020-117263GB-100) and XAI-Disinfodemics (PLEC2021-007681) grants, by Comunidad Aut\'{o}noma de Madrid under S2018/TCS-4566 grant, by European Comission under IBERIFIER - Iberian Digital Media Research and Fact-Checking Hub (2020-EU-IA-0252), by Digital Future Society (Mobile World Capital Barcelona), under the project DisTrack - Tracking disinformation in Online Social Networks through Deep Natural Language Processing, and by "Convenio Plurianual with the Universidad Politécnica de Madrid in the actuation line of \textit{Programa de Excelencia para el Profesorado Universitario}".

\bibliographystyle{unsrtnat}

\bibliography{references}  


\end{document}